\documentclass[pdflatex,sn-nature]{sn-jnl}

\usepackage{graphicx}%
\usepackage{multirow}%
\usepackage{amsmath,amssymb,amsfonts}%
\usepackage{amsthm}%
\usepackage{mathrsfs}%
\usepackage[title]{appendix}%
\usepackage{xcolor}%
\usepackage{textcomp}%
\usepackage{manyfoot}%
\usepackage{booktabs}%
\usepackage{algorithm}%
\usepackage{algorithmicx}%
\usepackage{algpseudocode}%
\usepackage{listings}%
\usepackage{subcaption}
\usepackage{lineno}

\theoremstyle{thmstyleone}%
\theoremstyle{thmstyletwo}%

\theoremstyle{thmstylethree}%

\begin{document}
\title[Article Title]{Neural Responses to Affective Sentences Reveal Signatures of Depression}

\author*[1]{\fnm{Aditya} \sur{Kommineni}}\email{akommine@usc.edu}
\author[1,2]{\fnm{Woojae} \sur{Jeong}}
\author[3]{\fnm{Kleanthis} \sur{Avramidis}}
\author[4]{\fnm{Colin} \sur{McDaniel}}
\author[4]{\fnm{Myzelle} \sur{Hughes}}
\author[5]{\fnm{Thomas} \sur{McGee}}

\author[6]{\fnm{Elsi} \sur{Kaiser}}
\author[3]{\fnm{Kristina} \sur{Lerman}}
\author[5]{\fnm{Idan A.} \sur{Blank}}
\author[6]{\fnm{Dani} \sur{Byrd}}
\author[4]{\fnm{Assal} \sur{Habibi}}
\author[4,7]{\fnm{B. Rael} \sur{Cahn}}
\author[1]{\fnm{Sudarsana} \sur{Kadiri}}
\author[1]{\fnm{Takfarinas} \sur{Medani}}
\author[1]{\fnm{Richard M.} \sur{Leahy}}

\author[1,3,6,8]{\fnm{Shrikanth} \sur{Narayanan}}

\affil[1]{\orgdiv{Ming Hsieh Department of Electrical and Computer Engineering}, \orgname{University of Southern California}}

\affil[2]{\orgdiv{Alfred E. Mann Department of Biomedical Engineering}, \orgname{University of Southern California}}

\affil[3]{\orgdiv{Thomas Lord Department of Computer Science}, \orgname{University of Southern California}}

\affil[4]{\orgdiv{Brain and Creativity Institute}, \orgname{University of Southern California}}

\affil[5]{\orgdiv{Department of Psychology}, \orgname{University of California Los Angeles}}

\affil[6]{\orgdiv{Department of Linguistics}, \orgname{University of Southern California}}

\affil[7]{\orgdiv{Department of Psychiatry and Behavioral Sciences}, \orgname{University of Southern California}}

\affil[8]{\orgdiv{Department of Psychology}, \orgname{University of Southern California}}

\abstract{Major Depressive Disorder (MDD) is a highly prevalent mental health condition, and a deeper understanding of its neurocognitive foundations is essential for identifying how core functions such as emotional and self-referential processing are affected. We investigate how depression alters the temporal dynamics of emotional processing by measuring neural responses to self-referential affective sentences using surface electroencephalography (EEG) in healthy and depressed individuals. Our results reveal significant group-level differences in neural activity during sentence viewing, suggesting disrupted integration of emotional and self-referential information in depression. Deep learning model trained on these responses achieves an area under the receiver operating curve (AUC) of 0.707 in distinguishing healthy from depressed participants, and 0.624 in differentiating depressed subgroups with and without suicidal ideation. Spatial ablations highlight anterior electrodes associated with semantic and affective processing as key contributors. These findings suggest stable, stimulus-driven neural signatures of depression that may inform future diagnostic tools.}
\maketitle
\section{Introduction}
Depression is one of the most prevalent mental health disorders worldwide, with estimates indicating that around 5\% of the worlds' adult population \cite{depression2017other, goodwin2022trends} suffers from this condition. 
The primary methods for screening and monitoring depression rely on self-reported questionnaires, such as the Patient Health Questionnaire (PHQ-9) \cite{kroenke2001phq}, Beck's Depression Inventory (BDI) \cite{beck1961beck} and Hamilton Depression Ratings Scale (HDRS) \cite{hamilton1960rating}. 
While these questionnaires are effective to varying degrees at screening patients for depression, they provide only limited information about the affected underlying neuro-cognitive processes in individuals, limiting the ability to personalize treatments.
Given the heterogeneity of depressive symptomatology across patient populations \cite{musliner2016heterogeneity, buch2021dissecting}, it is crucial to elucidate the underlying neurophysiological mechanisms to support the development of more effective and individualized procedures for screening, monitoring, and treatment.
\par 
Prior functional imaging studies have identified increased activity in anterior cingulate cortex (especially the subgenual anterior cingulate) during presentation of emotional stimuli,  altered connectivity in prefrontal cortical areas, and default mode network as potential differentiating markers in depressed participants \cite{schlosser2008fronto, dichter2009affective, wang2012systematic, pilmeyer2022functional, piani2022sustained, mohammadi2023brain}. Additionally, data from resting-state functional Magnetic Resonance Imaging (fMRI) have enabled successful classification of depressed participants \cite{rosa2015sparse, ramasubbu2016accuracy, gallo2023functional}, underscoring the promise for identifying biomarkers for depression.  
The scalability, low cost, and high temporal resolution of electroencephalography (EEG) make it a more practical modality for studying neural characteristics in depression and positioning it as a promising tool for large-scale screening in clinical settings \cite{de2019depression, simmatis2023technical}.
While there has been considerable interest in exploring resting-state differences between healthy and depressed populations \cite{koo2017current, vcukic2020classification}, task-based responses offer potentially greater specificity \cite{pelosi2000working, segrave2010upper, galkin2020impairments, keren2018reward, jiang2021enhancing} to decipher specific disruptions (e.g., working memory, reward processing, emotional processing) in clinical populations with mental health disorders.
\par 
Emotional processing has been observed to be a core component impacted in major depressive disorder (MDD) \cite{stuhrmann2011facial, carballedo2011functional, godlewska2016early, li2021abnormal}. 
While some studies have found negative processing bias in a scrambled sentence task to be predictive of subsequent depressive symptoms \cite{rude2002negative}, others report reduced neural activity to processing positive emotional information in a working memory task \cite{shestyuk2005reduced}.
Subsequent studies have shown impaired semantic sentence processing in depressed individuals \cite{iakimova2009behavioral} but offer conflicting insights related to sentence processing effects, with some studies \cite{klumpp2010semantic} noting no significant differences in N400 Event-related Potentials (ERP) response during a sentence processing task in healthy vs. depressed participants, and others \cite{kiang2017abnormal} observing reduced N400 amplitudes in depressed participants for a self-referential phrase ending in negative adjectives.
This warrants further investigation into whether healthy and depressed individuals exhibit distinct neural responses during the reading of affectively salient sentences, and whether these differences can serve as a reliable tool for screening.
\par 
The present study employs a self-referential sentence response paradigm to investigate temporal dynamics of affective self-referential sentences in healthy and depressed populations, along with quantifying the ability to distinguish mental health groups through the recorded neural signals. 
While prior studies using self-referential sentence response tasks have identified differences in ERP \cite{shestyuk2010automatic, auerbach2015self, benau2019increased}, they have not assessed the feasibility of using this experimental design to distinguish healthy and depressed individuals in a classification paradigm. 
Our objectives are twofold: (1) to examine clinical versus healthy group-level differences in the temporal profile of neural responses to the sentence response task through Multivariate Pattern Analysis; and (2) to evaluate the feasibility of using these neural responses to classify depression using a deep learning framework.
The temporal dynamics reveal baseline differences between healthy and depressed participants, agnostic to the semantic nature of the sentence. This indicates a difference in cognitive processing during sentence viewing, presented in a one-word-at-a time reading task.
Additionally both the temporal dynamics and deep learning experiments exhibit better distinguishability between healthy and depressed groups for positive sentiment sentences compared to negative sentiment sentences.
A deep learning model to classify healthy and depressed participants achieves an AUC (Area Under the Receiver Operating Characteristic curve) of 0.707, highlighting the potential of the experimental paradigm and modeling approach. 
Additionally, attempts to classify subgroups within depressed participants —depressed non-suicidal versus depressed suicidal —yield an AUC of 0.624, while promising, indicate open challenges when differentiating subgroups within depressed participants.

\section{Results}
As part of the sentence viewing task, participants viewed sentences presented in a one-word-at-a-time manner and were prompted to indicate whether they agreed or disagreed with the sentence after its presentation, while EEG was simultaneously recorded.
The task comprised 160 unique sentences with 80 minimal pairs of sentences, wherein each pair differed in the last word only. 
Within each pair, typically one sentence was designed to be congruent with the beliefs of depressed participants (generally negative sentiment) whereas the other being congruent with the beliefs of healthy participants (generally neutral or positive sentiment). 
This enables investigating whether neural responses reveal significant group differences between healthy and depressed participants based on the semantic nature of the sentence.
\subsection{Temporal Dynamics of Sentence Response Task}
For this analysis, within-subject Multivariate Pattern Analysis (MVPA) \cite{haxby2012multivariate, cichy2017multivariate} was adapted to a between-group framework.
This enabled investigating the temporal windows during which neural responses of healthy controls (C) differ from those of depressed non-suicidal (D) and depressed suicidal (S) participants.
To identify the types of affective stimuli that reveal larger differences in neural responses, group-level decoding was performed for the trials grouped based on their affective sentiment: (1) all sentences, (2) positive sentiment sentences, (3) neutral sentiment sentences, (4) negative sentiment sentences and (5) contrasting sentences (positive sentiment sentences - negative sentiment sentences; refer Sec.~\ref{sec:stim_char}).
The trials corresponding to sentence presentation were epoched from 200ms before to 900ms after onset of the last word in the sentence.
For each aforementioned sentence group, one ERP per subject was computed by averaging all trials corresponding to the particular sentence group. 
These ERPs were then input to a time-resolved classification pipeline, where a separate classifier is trained at each time point.  
The classification pipeline consists of a logistic regression classifier with L1-regularization.
The dataset comprises participants from two broad mental health groups, healthy Controls (C) and depressed, which is further subdivided into depressed non-suicidal (D) and depressed suicidal (S) groups.
In this analysis, group-level decoding is performed between healthy (C) and depressed participants (D, S), with AUC being reported as the evaluation metric.
All experiments are conducted in a between-subjects manner, with the average AUC being reported for a Leave-One-Subject-Out Cross-Validation (LOSOCV) setting across 10 random seeds.
\par
\begin{figure}
    \centering
    \includegraphics[width=0.8\linewidth]{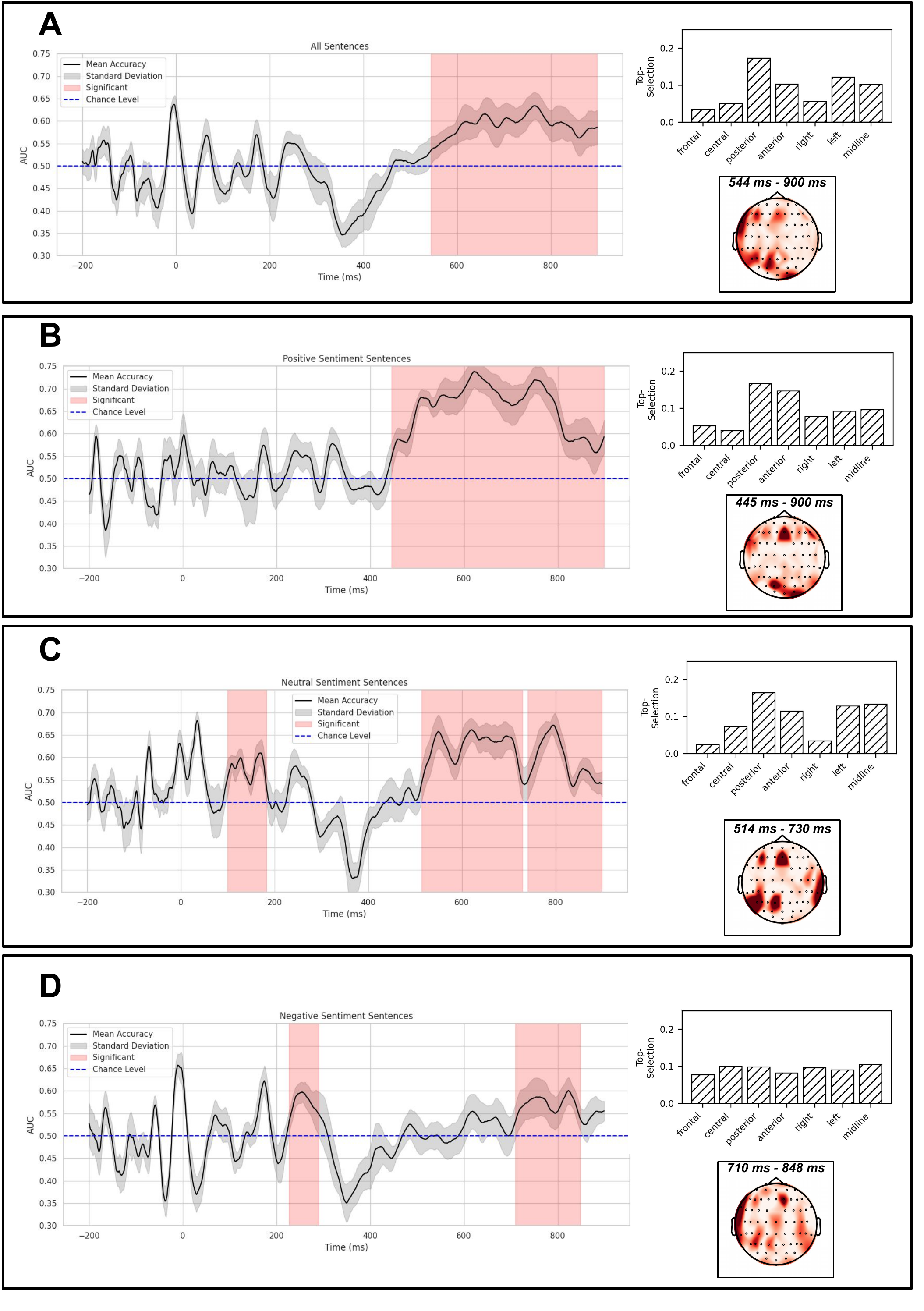}
    \caption{Temporal plots for per-timestep average AUC for Control versus Depressed non-suicidal \& Depressed suicidal [C $vs$ D,S] classification in a LOSOCV setting across 10 different seeds for (A) All sentences (B) Positive sentiment sentences (C) Neutral sentiment sentences and (D) Negative sentiment sentences. Shaded regions correspond to clusters wherein the performance were found to be statistically significantly above chance from a cluster based permutation test (p$\leq$0.05). The histogram plot and channel importance map of the largest cluster in each setting are reported. Channel Importance Map: Each value represents the proportion of significant timepoints during which the corresponding electrode was assigned a non-zero weight in L1-regularized logistic regression, indicating its contribution to group classification. Darker regions in the weight proportion map indicate the specific region to have higher contribution to C $vs$ D,S classification.}
    \label{fig:sen-sent-decoding}
\end{figure}

 
While prior studies have reported group differences in processing positive- or negative-sentiment sentences \cite{rude2002negative, shestyuk2005reduced, iakimova2009behavioral, klumpp2010semantic, kiang2017abnormal}, here we observe significant group differences between healthy controls and depressed participants for \textit{all sentences} group (positive + neutral + negative sentiment sentences) between (544 ms, 900 ms) from onset of the last word presentation in the sentence as seen in Fig.~\ref{fig:sen-sent-decoding}A.
This indicates a baseline difference in neural signals between healthy and depressed participants, potentially arising from differences in underlying cognitive processing during integration of meanings over time and context, and decision making~\cite{zhou_posterior_2019, binder2009semantic, valdebenito-oyarzo_parietal_2024,gold2007neural}, as the region of significance (544 ms to 900 ms) corresponds to the interval immediately preceding response prompt in which participants indicate whether they agree or disagree with the sentence.
Channel importance maps in the statistically significant regions indicate group differences arising from posterior regions of the scalp with bias towards left hemisphere.
\par
\begin{figure}
    \centering
    \includegraphics[width=1\linewidth]{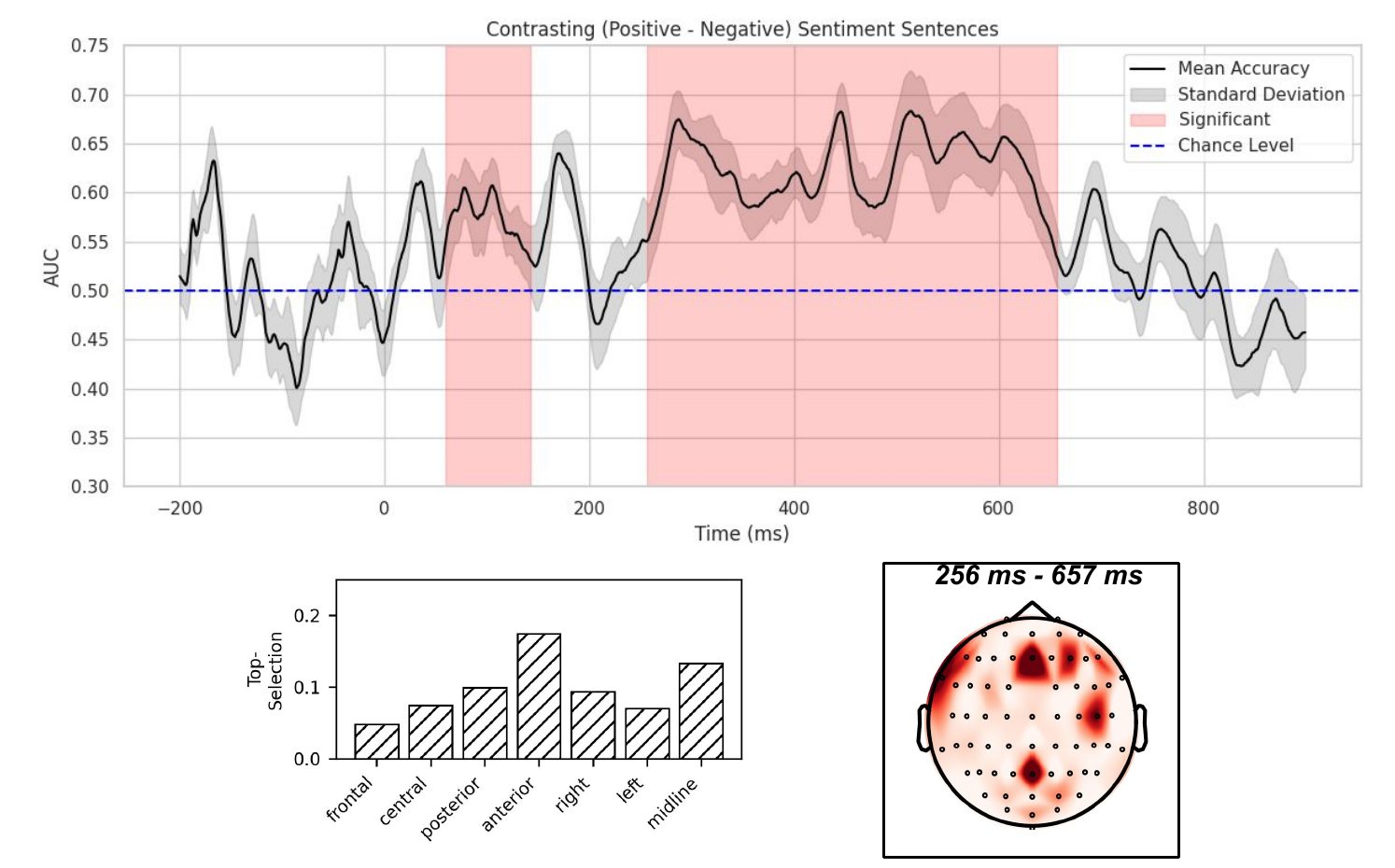}
    \caption{Decoding plot for Control versus Depressed non-suicidal \& Depressed suicidal [C $vs$ D,S] classification with contrasting trials (positive sentiment - negative sentiment). Timesteps with statistically significant performance above chance are shaded. Channel Importance map indicates anterior electrodes to provide differentiating responses between healthy and depressed participants.}
    \label{fig:sen-sent-decoding-contrasting}
\end{figure}

 
When performing group decoding with sentences grouped based on sentiment, a significant region of decodability is observed around 500ms after onset of the final word for positive (Fig.~\ref{fig:sen-sent-decoding}B), neutral (Fig.~\ref{fig:sen-sent-decoding}C), and negative (Fig.~\ref{fig:sen-sent-decoding}D) sentiment sentence groupings.
In addition to the late component, neutral and negative sentences show an early component of significant decodability --from 100 ms to 183 ms and 226 ms to 290 ms respectively--as shown in Fig.~\ref{fig:sen-sent-decoding}C,~\ref{fig:sen-sent-decoding}D. 
For neutral and positive sentences, anterior and posterior channels show a greater proportion of importance during classification. 
This could indicate differences in late positive potential, which has been found in prior studies to originate as a result of differences in predictability of words, with distinct components in the anterior and posterior regions \cite{van2012prediction, delong2014predictability}.
Such anterior and posterior bias in channel importance map is not observed for negative sentences.
Additionally, peak AUC value and the statistically significant cluster sizes for positive and neutral sentences are noticeably higher compared to negative sentences, suggesting that group differences in ERPs are less pronounced for negative sentence processing.
When contrasting (subtracting positive and negative sentiment sentence ERPs), midline electrodes in anterior channels which are primarily responsible for emotional processing provide the highest distinguishability (Fig.~\ref{fig:sen-sent-decoding-contrasting}).
The temporal cluster with statistically significant decodability is from (256 ms to 657 ms), revealing earlier distinct responses compared to individual groups of sentences.

\subsection{Trial Grouping Based Depression Classifier}
\begin{table}
\centering
\caption{AUC and 95\% Confidence Intervals (CI) for EEG classification across sentence grouping conditions and trial types. Statistical significance was determined using permutation tests against the null hypothesis of AUC = 0.5, with * indicating $p<0.05$ and ** indicating $p\leq0.001$. For the baseline rows, \textit{All Sentences} corresponds to grouping all sentences together, while \textit{Random Contrasting} splits the sentences randomly into two groups.}
\begin{tabular}{llcc}
\toprule
\textbf{Condition} & \textbf{Trial Type} & \textbf{C $vs$ DS} & \textbf{D $vs$ S} \\
\midrule

\multirow{3}{*}{Sentence Sentiment} 
    & Positive     & 0.676 [0.63, 0.72]**  & 0.517 [0.46, 0.57] \\
    & Negative     & 0.549 [0.50, 0.59]*   & 0.594 [0.54, 0.65]** \\
    & Contrasting  & \textbf{0.707 [0.66, 0.75]}**  & 0.598 [0.54, 0.65]** \\
\midrule

\multirow{3}{*}{Last Word Valence} 
    & Positive     & 0.586 [0.54, 0.63]**  & 0.559 [0.51, 0.61]* \\
    & Negative     & 0.615 [0.57, 0.66]**  & 0.553 [0.50, 0.61]* \\
    & Contrasting  & 0.604 [0.55, 0.64]**  & 0.539 [0.486, 0.591] \\
\midrule

\multirow{3}{*}{Response Type} 
    & Agree        & 0.636 [0.59, 0.68]**  & 0.551 [0.50, 0.61]* \\
    & Disagree     & 0.588 [0.55, 0.63]**  & 0.542 [0.49, 0.60] \\
    & Contrasting  & 0.688 [0.64, 0.73]**  & 0.606 [0.55, 0.65]** \\
\midrule

\multirow{3}{*}{Response Time} 
    & Slow         & 0.625 [0.58, 0.66]**  & 0.526 [0.47, 0.58] \\
    & Fast         & 0.609 [0.56, 0.65]**  & 0.576 [0.52, 0.63]** \\
    & Contrasting  & 0.650 [0.60, 0.69]**  & \textbf{0.624 [0.57, 0.67]}** \\
\midrule

\multirow{2}{*}{Baseline} 
    & All Sentences         & 0.587 [0.54, 0.63]**  &  0.554 [0.50, 0.61]* \\
    & Random Contrasting    & 0.624 [0.58, 0.66]**  & 0.579 [0.53, 0.63]** \\
\bottomrule
\end{tabular}
\label{tab:auc_results}
\end{table}

While group decoding analyses primarily emphasize the temporal dynamics of neural signals, they do not allow for the models to leverage the spatiotemporal information of EEG responses.
Hence, deep learning models were trained to classify mental health groups (Healthy (C), Depressed non-suicidal (D) and Depressed suicidal (S)).
The EEG signals from 200 ms before to 900 ms after the onset of the final word of the sentence were used as input to the model.
Classification results using the aforementioned model are presented in Table~\ref{tab:auc_results}, across 4 broad categories of sentence grouping (refer Sec.~\ref{sec:stim_char}): (1) Sentence sentiment, (2) last word valence, (3) response type (agreement versus disagreement) and (4) response time of participant. 
The baseline category contains two conditions, with \textit{all sentences} grouping the sentence set into a single group, while \textit{random contrasting} divides the sentences into two groups at random and then generates inputs by subtracting the responses from the two groups.
Models are trained for two classification tasks, (1) C $vs$ (D, S) for distinguishing healthy and depressed participants, and (2) D $vs$ S, aiming to disentangle depressed non-suicidal (D) from depressed suicidal (S) participants. 
\par 
Majority of sentence grouping conditions result in classifiers with statistically greater than chance performance for C $vs$ D,S classification, with sentence-sentiment contrasting models reaching an AUC of \textit{0.707 [95\% CI = (0.66, 0.75)]}.
Depressed sub-group classification models (D $vs$ S) resulted in more modest performance with response time contrasting models providing an AUC of \textit{0.624 [95\% CI =  (0.57, 0.67)])}.
When compared to baseline model (AUC=0.587), training on positive sentiment sentences (AUC=0.676) or response type agreement sentences (AUC=0.636) conditions provide performance gains for C $vs$ D,S classification, while for D $vs$ S classification negative sentiment sentences (AUC=0.594) provide marginally better performance over baseline (AUC=0.554) indicating the utility of grouping sentences based on the linguistic and response characteristics of the stimuli.
It is worth noting that, unlike a random-chance or majority-voting baseline, the ``all sentences" baseline model captures neural signal differences between mental health groups in the sentence response task without relying on prior knowledge of stimulus type. 
Hence, any statistically significant performance over this method indicates the utility of trial grouping.
\par 
Contrasting involves subtracting the trial types within a category of sentence grouping, e.g., for sentence sentiment the positive and negative sentiment trials were subtracted. 
For contrasting models of sentence sentiment, response type and response-time grouping, performance gains are observed relative to their respective individual trial types in both for C $vs$ D,S and D $vs$ S classification tasks. 
This could be due to complementary information when subtracting opposite categories of a condition (positive sentiment $vs$ negative sentiment or agree $vs$ disagree). 
Within the individual trial type of sentence grouping, positive sentiment sentences provide the best distinguishability for C $vs$ D,S classification.  This is consistent with similar findings in prior work wherein reduced activity in a depressed population was observed when processing positive sentiment words during a working memory task, but no such differences were observed for negative sentences \cite{shestyuk2005reduced}.
Additionally, this observation is in accordance with the group decoding results wherein positive sentiment sentences provided higher AUC and larger statistically significant cluster sizes over negative sentiment sentences.
Interestingly, while positive sentiment sentences provide better distinguishability for C $vs$ D,S classification, negative sentiment sentences show better performance for D $vs$ S indicating the utility of both positive and negative sentiment sentences.
\par
With regard to comparison between sentence sentiment-based and last-word valence-based groupings, a consistent trend of greater discriminability is observed for sentence sentiment.
This could be owing to task design emphasizing on participants requiring to respond to the displayed sentence, resulting in the sentence sentiment being more effective at eliciting consistent differences. 
When comparing response type based grouping and sentence sentiment based, no consistent performance gains are observed for one condition type. However the response type grouping requires explicit feedback from participants whereas the sentence sentiment is based entirely on the semantic characteristics of the sentence stimuli. This could be promising for future experimental approaches that do not necessitate an explicit feedback from the participants. 
\subsection{Interpretability Analysis}
\par
\begin{figure}
  \centering
  \begin{subfigure}{0.49\textwidth}
    \includegraphics[width=\textwidth]{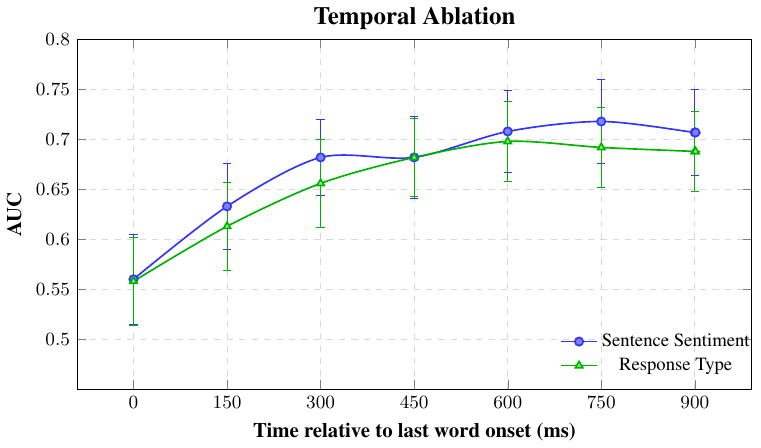}
    \label{fig:temporal-performace}
  \end{subfigure}
  \hfill
  \begin{subfigure}{0.49\textwidth}
    \includegraphics[width=\textwidth]{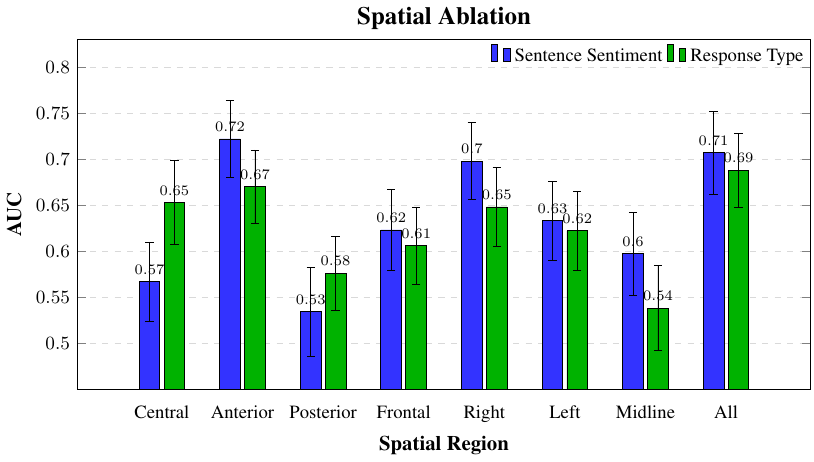}
    \label{fig:spatial-performace}
  \end{subfigure}
  \caption{Performance when varying the temporal and channel inputs to the model. \texttt{LEFT:} Temporal performance with input signal as (-200 ms to X ms); X $\in$ {0, 150, 300, 450, 600, 750, 900}. Performance increases as the input size increases and saturates around 600 ms. \texttt{RIGHT:} Performance variation when varying the input channels across different electrode regions. Anterior channels show better performance across both conditions, while central channels show importance in the response type model.}
  \label{fig:temp-spatial}
\end{figure}

 \textbf{Temporal \& Spatial Regions of Discriminability} provide interpretable information highlighting the time regions and electrodes of significance for distinction between the groups. 
To this end, ablations on the two best performing models for C $vs$ D,S classification models (sentence sentiment contrasting, response type contrasting) are plotted in Fig.~\ref{fig:temp-spatial}.
For temporal ablation, the input to the deep learning model was constrained to (-200 ms to X ms) with X $\in$ \{0, 150, 300, 450, 600, 750, 900\}. 
The 900 ms input signal contains the entire duration of last word presentation and the inter-word interval of 300 ms before the subject responds to the sentence; whereas 0 ms input signal includes 200 ms prior to the presentation of the last word, which corresponds to the inter-word interval of the penultimate in the sentence.
Both models show an expected increase in AUC magnitude as the length of the input signal increases.
For the response type model, a saturation in performance is observed around 600 ms, whereas an increasing performance trend is seen in the sentence sentiment model up to 750 ms.
In spatial localization ablation, the sentence sentiment contrasting model shows similarities to decoding experiments (sentence sentiment contrasting decoding experiment wherein channel importance map [refer to Fig.~\ref{fig:sen-sent-decoding-contrasting}] indicates differences arising from anterior channels) with anterior channels providing better classification. For the response type contrasting model, central and anterior channels provide better differentiation.
These spatial ablation results provide promising directions for considering the reduction in the number of electrodes used in data collection for future studies.

\begin{figure}
    \centering
    \includegraphics[width=\linewidth]{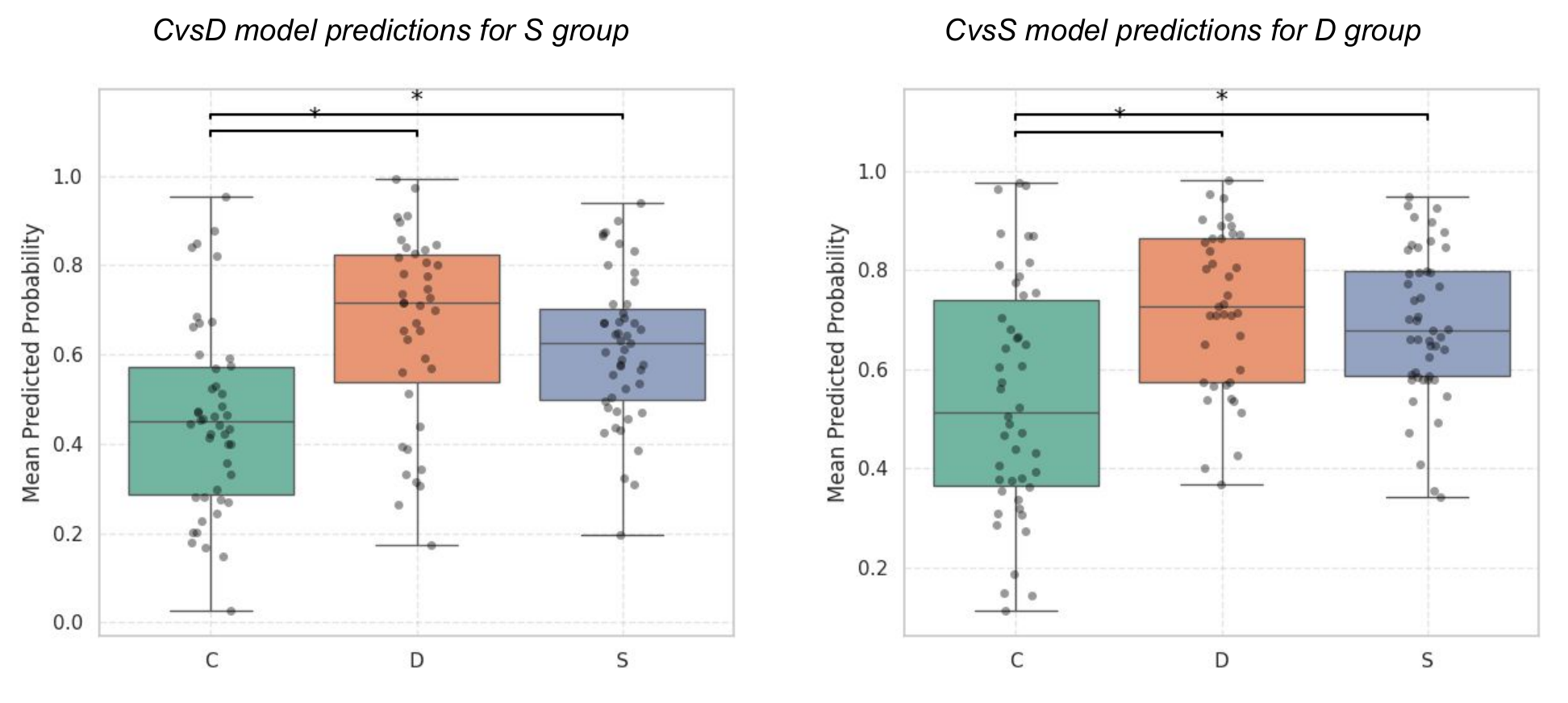}
    \caption{Boxplot of prediction probabilities of models across the three groups of participants. \texttt{Left}: Model trained for C $vs$ D; \texttt{Right}: Model trained for C $vs$ S. At testing time, for both models all the three groups (C, D and S) are predicted. Statistical tests indicate the model probabilities to be statistically different (p$<$0.05 after bonferroni correction for multiple t-tests) between C and D, C and S for both the models indicating generalization of predictions for D and S groups.}
    \label{fig:unseen-group}
\end{figure}
\par \textbf{Generalizability to Unseen Subgroup} is an essential requirement for clinical diagnostic models that require translation across heterogeneous patient populations. 
In the context of depression, which encompasses multiple subtypes and symptom profiles this generalizability is critical for developing tools that are robust and widely applicable. 
Such models support the development of scalable, front-line screening tools that do not require fine-tuning for each clinical subtype, thereby enhancing real-world applicability.
To evaluate this, we assessed how well models trained on one clinical subgroup generalized to another. Specifically, we trained separate models for (C vs D) and (C vs S) classifications, where either depressed suicidal or depressed non-suicidal individuals were excluded from the training set, respectively. 
The performance of these models was then tested on the held-out subgroup by examining the average probability that the model assigned to depressed class labels for the unseen group. 
Fig.~\ref{fig:unseen-group} reports the average depressed probabilities predicted by the model for test participants in the three groups, with the plot on the left trained for (C $vs$ D) and the model on right is trained on (C $vs$ S). 
The (C vs S) model predicted an average depressed probability of 0.72 for depressed non-suicidal participants (D), while the (C vs D) model assigned an average depressed probability of 0.61 to depressed suicidal participants (S). 
Importantly, for both models, the predicted probabilities for the unseen depressed subgroup were significantly higher than those for healthy participants, yet no significant differences were found between the two depressed subgroups, suggesting consistency in the neural signatures of depression across subtypes.
Further, the (C vs D) and (C vs S) models achieved AUCs of 0.72 [95\% CI: 0.68–0.76] and 0.69 [95\% CI: 0.64–0.73], respectively, when evaluated on the full (C vs D,S) task. 
These results support the robustness of the models to missing clinical subgroups during training, underscoring their potential for broader deployment in diverse patient populations where detailed subtype labeling may not be available. 
\par
\begin{figure}
  \centering
  \begin{subfigure}{0.49\textwidth}
    \includegraphics[width=\textwidth]{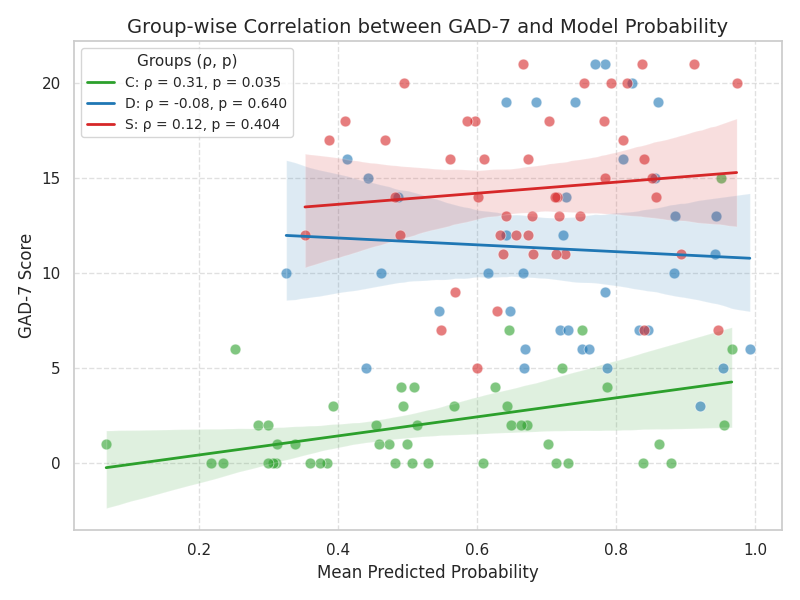}
    \label{fig:gad7_corr}
  \end{subfigure}
  \hfill
  \begin{subfigure}{0.49\textwidth}
    \includegraphics[width=\textwidth]{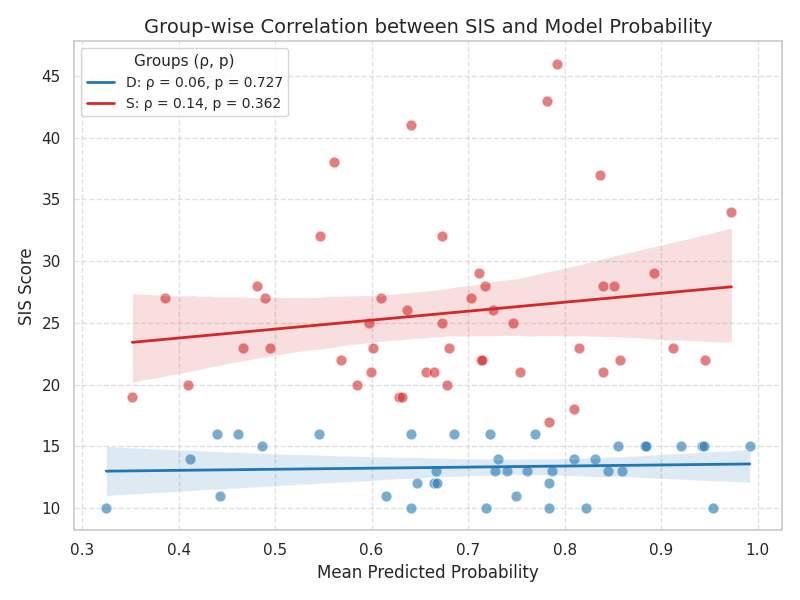}
    \label{fig:sis_corr}
  \end{subfigure}
  \caption{Correlations of contrasting sentence sentiment model probabilities for C $vs$ D, S task with \texttt{Left:} GAD-7 and \texttt{Right:} SIS questionnaires. Spearman correlation is reported per group and the p-values test against a null distribution of zero correlation.}
  \label{fig:correlation-plot}
\end{figure}

 \textbf{Correlations of Predictions to Questionnaires:}
PHQ-9 scores of the participants are recorded twice, first during the recruitment screening phase and second on day of EEG recording. 
A higher correlation between model depression probabilities and PHQ-9 score on the day of the EEG recording (0.327) was observed as compared to the PHQ-9 from the earlier screening (0.243), and this difference is found to be statistically significantly based on a permutation test (p=0.003). 
While a positive correlation with PHQ-9 score across groups is expected owing to the groups being defined based on the score, a higher score with the PHQ-9 on the day of the screening could indicate that the model depression probabilities are legitimately capturing within group effects.
In addition to the PHQ-9 questionnaire, SIS \cite{rudd1989prevalence} and GAD-7 \cite{williams2014gad} questionnaires are completed by the participants during screening and on the day of recording respectively.
Plots for within-group correlations for SIS and GAD-7 scores can be seen in Fig.~\ref{fig:correlation-plot}.
Positive correlation between model depression probability and questionnaire score is observed for controls and suicidal participants in GAD-7 score and for suicidal participants in SIS score. 
While the trends show positive correlation, these effects are not found to be statistically significant. 
However, the ability of the models to assign higher depressed probability to participants with higher questionnaire scores without explicitly providing this information could indicate that the models are implicitly  able to model symptom severity.

\section{Discussion}
In this study, we investigate whether neural responses to emotional self-referential stimuli can serve as reliable biomarkers for distinguishing between healthy and depressed individuals. 
While prior research has demonstrated group-level differences in event-related potentials (ERPs) to emotional stimuli in depression \cite{rude2002negative, shestyuk2005reduced, klumpp2010semantic}, these findings have not yet been systematically evaluated for their utility in developing objective, data-driven screening tools.
Our study makes a twofold contribution: first, by examining the temporal dynamics of neural responses to affective self-referential sentences using multivariate pattern analysis (MVPA); and second, by evaluating the feasibility of deep learning-based classification approaches for potential use in practical screening applications.
MVPA results reveal that group differences in neural responses likely reflect a combination of emotional processing and decision-related cognitive mechanisms. 
We observe baseline group differences in response to self-referential sentences irrespective of the sentiment of the sentence. 
These differences predominantly emerge at latencies $\ge$ 500 ms, coinciding with the conclusion of the sentence presentation and immediately preceding the behavioral prompt (agree/disagree). 
This timing suggests that the observed group differences may be driven by decision-making processes and associated higher-order cognitive functions engaged during evaluative judgment.
On the other hand, for group decoding with contrasting (positive - negative sentiment sentence) condition, we identify earlier neural differences between groups in the 250–650 ms window. 
These differences are localized in anterior scalp regions, consistent with known correlates of emotional and affective processing \cite{citron2012neural}. 
Crucially, the absence of significant group differences beyond 650 ms in this condition implies that later-stage neural differences are not modulated by sentence sentiment. 
Instead, they appear to reflect general decision making processes which are common across stimuli. 
The early group differences, however, likely reflect altered attentional or emotional responses triggered by the presentation of the affect-critical final word in the sentence, aligning with prior work on affective sentence processing \cite{friederici2011brain}.
Together, these findings suggest temporally and functionally dissociable neural markers of depression: early activity linked to emotion-specific stimulus processing and later activity related to decision-making. 
\par 
The results of our deep learning models underscore the potential of EEG as a viable modality for identifying neurophysiological biomarkers of depression. 
Achieving an AUC of 0.707 in distinguishing healthy individuals from depressed participants demonstrates that temporally dynamic neural responses to self-referential emotional stimuli encode discriminative patterns relevant to clinical status. 
While the sensitivity of the model is relatively high (73\%), the lower specificity (60\%) suggests that a subset of healthy individuals are misclassified as depressed. 
This misattribution may in part stem from the class imbalance in the dataset, where the proportion of depressed participants was larger. 
In addition to the overall classification performance, the spatial and temporal ablation analyses offer important practical insights. 
Specifically, they suggest the possibility of reducing the number of electrodes required without substantially compromising model accuracy. 
This opens the door to more portable and clinically viable EEG configurations that minimize setup time and participant burden.
\par 
\begin{figure}
    \centering
    \includegraphics[width=\linewidth]{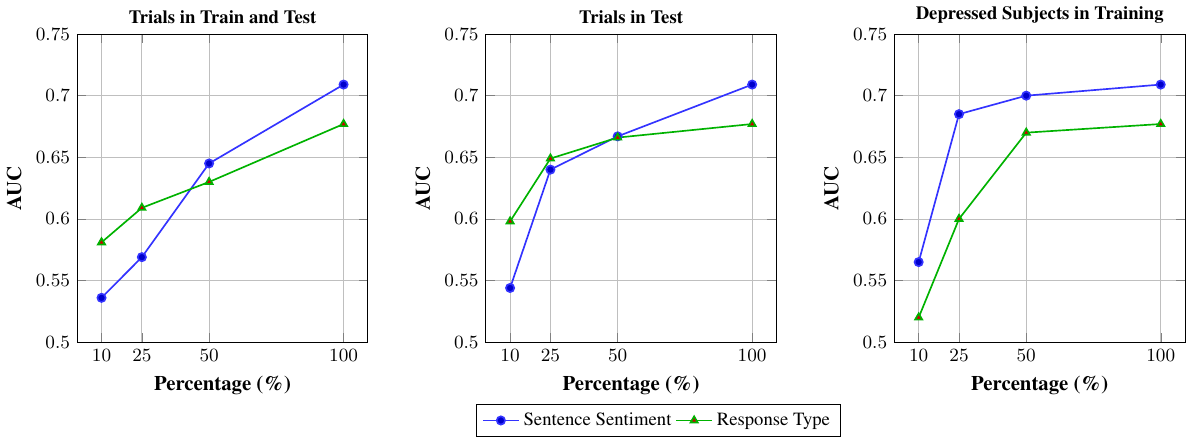}
    \caption{Ablations on number of trials presented per subject and the number of depressed participants in training data for C $vs$ D,S task. \texttt{Left}: Reduction in trials in both train and test data. models show limited ability to learn with reduced trials in training data. \texttt{Middle}: When reducing the trials at testing time only, robustness to reduction in trials up to 75\% is observed for response type model. \texttt{Right}: Ablation reducing the percentage of depressed participants used. Sentence sentiment models indicate the ability to learn discriminating features under low depressed subject count.}
    \label{fig:data-ablations}
\end{figure}
A major consideration for the translational application of our approach, however, is the experimental duration. 
With a task length averaging 55 minutes, the current protocol may be too lengthy for routine clinical deployment. 
To evaluate whether data collection could be made more efficient, we examined the effect of reducing the number of trials per subject. 
Two ablation strategies were tested: one reducing trials during both training and testing, and another simulating a realistic deployment scenario in which a new subject is evaluated with a reduced trial set while training is conducted on full-length data. 
The results reveal that while reducing trial numbers across both training and testing leads to performance degradation (Fig.~\ref{fig:data-ablations} \texttt{Left}), limiting the number of trials only at test time (Fig.~\ref{fig:data-ablations} \texttt{Middle}) preserves model robustness even with up to a 75\% reduction. 
These findings suggest that broad stimulus variability during training is critical, but efficient screening at inference time may still be possible with fewer trials, offering a path toward shortened and more scalable EEG-based assessments.
Given the practical challenges in acquiring large-scale clinical EEG datasets especially from depressed populations, we also evaluated model performance under low-resource conditions. 
Remarkably, the sentence sentiment model maintained stable classification accuracy even when the number of depressed participants in training was reduced by up to 75\%, while the response type model tolerated reductions of up to 50\% with minimal impact as seen in Fig.~\ref{fig:data-ablations} \texttt{Right}. 
These findings indicate that the models can extract clinically meaningful features from relatively small samples of depressed individuals, which has important implications for the feasibility of training such models in real-world clinical contexts where participant recruitment is often limited.
\par 
To benchmark neural performance against a simpler behavioral baseline, we trained a logistic regression classifier using each participant’s response profile—namely, the number of agree versus disagree decisions to positive and negative sentences. 
This behavioral model achieved near-ceiling AUCs (0.94 for C vs D,S and 0.88 for D vs S), suggesting strong group-level differences in explicit evaluative responses. 
However, this approach may be less robust in practice, particularly in settings where participants are unwilling or unable to respond transparently due to stigma, discomfort, or motivational deficits.
An EEG-based model trained to predict the subject’s behavioral response achieved only chance-level performance (AUC = 0.552 ± 0.005), indicating that the neural classifier is not merely capturing response patterns but is instead leveraging distinct neural signatures associated with depression.
Together, these results highlight the promise of EEG-based models for depression screening and shed light on practical considerations regarding trial duration, data efficiency, and model robustness under real-world constraints. 
Future work may further refine these approaches for use in time-limited and resource-constrained clinical environments.
\section{Methods}
\subsection{Ethics Approval}
Our study was approved by the Institutional Review Board (IRB) of the University of Southern California (UP-23-00071). Personally identifiable information (PII) of the participants was de-identified for processing. 
Informed consent was obtained from all participants. 

\subsection{Recruitment and Task}
A cohort of 160 participants between 18 and 25 years old were recruited in the city of Los Angeles, California, for this study. 
Recruitment was performed using fliers on campuses of University of Southern California, University of California Los Angeles and Santa Monica College and through social media advertisements.
All participants were fluent in English and were required to complete a set of questionnaires to determine eligibility for the study. 
These forms included Patient Health Questionnaire (PHQ-9) \cite{kroenke2001phq}, and the Suicidal Ideation Scale (SIS) \cite{rudd1989prevalence} at the time of recruitment.
Participants who had: (1) current or previous diagnosis of neurological and psychiatric disorders including Schizophrenia, Bipolar Disorder, Epilepsy, Brain Cancer, and Stroke, (2) color blindness, (3) learned English after the age of 7, or (4) PHQ-9 scores in the range 5-9 were excluded from the study.
The participants who met the inclusion criteria of the study completed additional PHQ-9, GAD-7 \cite{williams2014gad} and TIPI \cite{gosling2003very} questionnaires on the day of the EEG recording. 
Participants were then grouped based on the questionnaire scores (recording during screening) as follows into \textit{Control (C, PHQ-9 $<$ 5)}, \textit{Depressed non-suicidal (D, PHQ-9 $>$ 9; SIS $\leq$ 16)} or \textit{Depressed suicidal (S, PHQ-9 $>$ 9; SIS $>$ 16)}. 
\par 
From the cohort of 160 participants, data corresponding to 13 participants with large variations in PHQ-9 scores from recruitment time to day-of (i.e., difference $\geq$5) and 1 participant who was unable to complete the experiment were excluded from the study.  
Of the remaining 146 participants, 49 were controls, 47 were depressed non-suicidal, and 50 were depressed suicidal.  
The demographic attributes for each group are shown in Table~\ref{tab:phq9_sis_cutoff_transposed}. 
\begin{table}
    \centering
    \caption{Demographics of the three mental health groups in the study. For Gender, M - Male, F - Female, O - Other. PHQ-9 and SIS thresholds were used to group participants into Controls (C), Depressed non-suicidal (D), and Depressed suicidal (S). Participants with PHQ-9 scores between 5–9 were excluded. Average age (years) with standard deviation is reported. Dominant hand indicates R = right, L = left.}
    \begin{tabular}{lccc}
    \toprule
         & C & D & S \\
    \midrule
         Gender (M/F/O) & 17/30/2 & 13/28/6 & 15/33/2 \\
         Total N & 49 & 47 & 50 \\
         Age (Mean ± SD) & 20.59 ± 1.92 & 21.15 ± 2.19 & 20.16 ± 1.83 \\
         Dominant Hand (R/L) & 45/4 & 39/7 & 49/1 \\
         PHQ-9 Criteria & $\leq$4 & $\geq$10 & $\geq$10 \\
         SIS Criteria & $\leq$16 & $\leq$16 & $>$16 \\
    \bottomrule
    \end{tabular}
    \label{tab:phq9_sis_cutoff_transposed}
\end{table}

\subsubsection{Sentence Task}
Participants took part in a sentence response task with simultaneous recording of EEG, eye-tracking and physiological signals.
In this analysis, we will focus on the EEG data. 
During the task, sentences presented to the participants were designed to contain 160 self-referential sentences with 80 minimal pairs such that each sentence in a pair differed in last word only. 
Within each pair, one sentence was designed such that its critical final word makes likely congruent to the beliefs and/or experiences of healthy participants (\textit{My mental health is not problematic} - neutral or positive sentiment sentences) and the other incongruent for healthy participants, and congruent for participants having compromised mental health (\textit{My mental health is not sound} - negative sentiment sentences).   That is, the sentences having positive affect were those likely to be congruent for control participants and those having negative affect were likely to be congruent for the clinical populations.  There were also sentences constructed in this manner that were biographical or demographic in content and designated neutral in sentiment. 
Sentences were 4 to 14 words long, and each was displayed twice in the protocol, resulting in 320 trials per subject; the sentences were displayed in a pseudo-random order.
\par
\begin{figure}
    \centering
    \includegraphics[width=\linewidth]{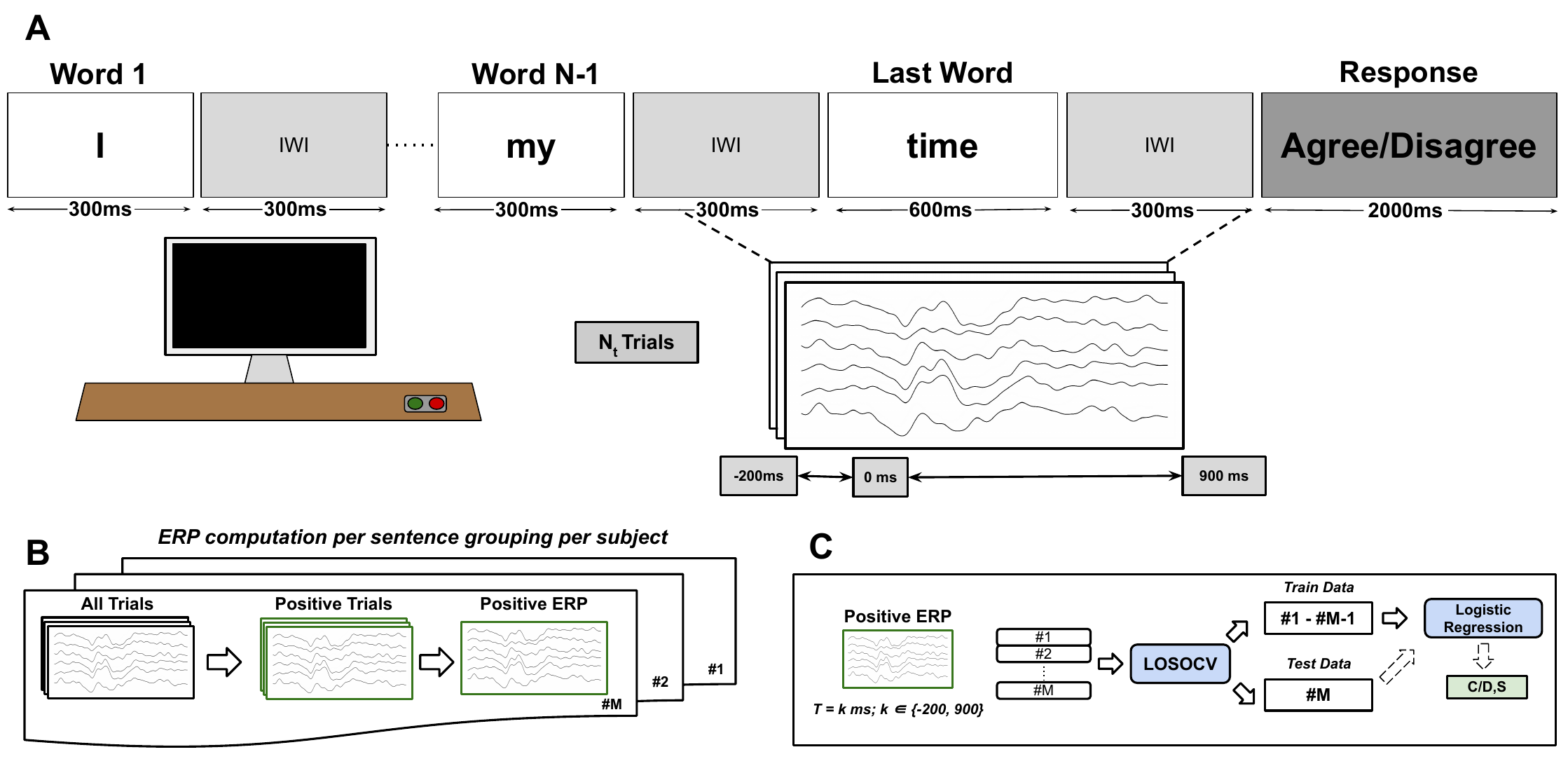}
    \caption{(\textbf{A}) Sentences are displayed one-word-at-a-time on a computer screen. As depicted in the image, each word is shown for 300 ms followed by an Inter Word Interval (IWI) of 300 ms blank screen. The last word in the sentence is displayed for 600 ms followed by 300 ms IWI following which the subject is prompted to respond with agree or disagree by pressing the respective button. Each sentence is shown twice in a pseudo random order amounting to 320 trials per subject. Parts \textbf{B}, \textbf{C} provide details of group decoding. (\textbf{B}) ERP of a specific sentence grouping (for example positive trials) was computed per subject. (\textbf{C}) After ERP computation, the participants data at every timestep are passed to a classification pipeline wherein Leave One Subject Out Cross Validation is performed with a logisitic regression classifier training per timestep per train test split.}
    \label{fig:methods_decoding}
\end{figure}

 
Each word of the sentence was displayed sequentially at the center of a screen for 300 ms followed by an inter-word-interval (IWI) of 300 ms (blank screen) with the last critical word of the sentence being displayed for 600 ms. 
At the end  of each sentence, the participants were given 2s to respond as to whether they agree or disagree with the sentence by clicking the respective button. 
Prior to presentation of the next sentence, a fixation cross at the center of the screen was displayed. 
Figure.~\ref{fig:methods_decoding}A provides a detailed pictorial representation of the recording setup and task design.   Groupings in the model and deep learning analysis were made both based on the sentiment of the sentence and on the individual subject response that confirmed or rejected this expected sentiment.  Valence of the final critical word was also considered as a relevant variable.  See further explication in Trial Grouping below.

For group decoding and deep learning analysis, each trial was epoched from 200 ms before the presentation of the final critical word of the sentence until 900 ms after the last word presentation. 
The time interval prior to onset of the last word was used to normalize the trial. 
The end of epoch 900 ms corresponds to the time at which the participants are prompted to respond to the trial. 
Hence, this epoching strategy minimized capturing any motor movement corresponding to the button-press response. 
\subsection{EEG Acquisition and Preprocessing}
EEG data was collected with a 64-channel active electrode system (actiCAP, Brain Products, GmbH) at a 1 kHz sampling rate. Impedances of the electrodes were lowered to under 30k$\Omega$ (often reached 5k$\Omega$-15k$\Omega$ range) by injecting an electrolyte gel (SuperVisc Gel) before the recording and were maintained under 30k$\Omega$ throughout the EEG recording. 
A semi-automated EEG preprocessing pipeline was implemented in MATLAB (Mathworks, Inc.) using Brainstorm \cite{tadel2011brainstorm}. EEGLab \cite{delorme2004eeglab} and FieldTrip \cite{oostenveld2011fieldtrip} toolboxes were used for bad/noisy segment detection. Powerline noise was removed with a 60 Hz IIR notch filter and the data was band pass filtered from 0.5 Hz to 80 Hz with an FIR filter.
Artifact Subspace Reconstruction (ASR) \cite{mullen2015real} routine was used for bad channel and bad time segment detection (channels with $<$0.8 correlation to neighboring channels, $\geq$ 4s flatline and ASR threshold 50 were marked as bad).
Bad channels were interpolated by averaging the neighboring channels within a 5cm distance. Bad time segments were temporarily removed from the data prior to Independent Component Analysis (ICA) \cite{makeig1995independent} for artifact removal. Following this, ICA was performed and artifactual components were manually removed (eye blinks, eye movements, electrocardiogram, and muscle movements). The remaining clean components were used to reconstruct the signal across the entire data.
A bad-time-segment detection routine was applied to the data after artifact removal.
The preprocessed data were re-referenced to the average of all channels, and the linear trend was removed.
The trials were epoched from -200 ms to 900 ms, i.e. from the stimulus onset, and were baseline corrected through Z-score transformation using the time window between -200 ms and stimulus onset.
During preprocessing, EEG data corresponding to 6 participants was found to be extremely noisy and was excluded from the analysis.
\subsection{Trial Grouping}
The set of 160 sentences displayed as a part of the sentence task vary across a variety of categories—sentiment of the sentence, valence of the last word, response time of the participants, agree/disagree to the sentence.
Differences in EEG responses across trials are investigated based on (1) Stimuli Characteristics and (2) Response Characteristics.

\subsubsection{Stimuli Characteristics}\label{sec:stim_char}
This category refers to grouping trials based on the linguistic characteristic of the sentences. 
Prior works indicate differences in ERP responses to both positive and negative sentiment sentences for healthy versus depressed participants, with some studies finding effects in positive sentiment sentences only \cite{shestyuk2005reduced}, while other studies find effects in negative sentiment sentences \cite{rude2002negative}.
While these study differences could be arising from variations in experimental design, one consistent observation is the presence of differences in affective sentence processing between healthy and depressed individuals.
Hence, in this study we choose two categories of grouping based on the stimuli characteristics (1) sentiment of the sentence (positive vs. negative) and (2) valence of the last word in the sentence (positive vs. negative).  (Recall that the valence of the final word [in the non-biographical sentences] was roughly balanced within the positive sentiment sentence and within the negative sentiment sentences.)
Additionally, we further hypothesize that rather than relying solely on an individual category of stimuli, such as positive and negative sentiment or valence, \textit{contrasting} these groups could provide better differentiability between the mental health groups.
Hence, the contrasting trials were computed by subtracting the groups of trials within a category (i.e., positive sentiment - negative sentiment sentences).
\subsubsection{Response Characteristics}
Prior works \cite{siegle2001pupillary} observed differences in reaction times between healthy and depressed participants when responding to self-relevant affective stimuli. 
It is of interest to explore whether these differences can be captured in neural signals prior to overt responses and whether these differences could result in robust group differentiating models.
We study response-based grouping along two dimensions, (1) Response type and (2) Response time. Response type grouping refers to grouping the trials based on whether the subject agrees with the sentence or disagrees with the sentence. Reaction time based grouping contains high and low groups with the high response time group containing the top 25\% reaction time trials of a subject and the low response time group the bottom 25\% reaction time trials.
\subsection{EEG Group Decoding}
Multi Variate Pattern Analysis (MVPA) \cite{haxby2012multivariate, cichy2017multivariate}, has been widely used in EEG and MEG studies to understand the temporal dynamics of visual processing \cite{cauchoix2014neural, hermann2022temporal}.
Here, ERPs of the participants with the sentences relevant to group of interest (positive sentiment sentences, agree sentences, or low reaction time sentences) were computed. 
At every timestep of the ERP, time steps within a window of 10ms were averaged and these inputs were passed to a classification pipeline for C $vs$ D,S classification in a Leave One Subject Out Cross Validation (LOSOCV) setting across 10 random seeds (refer to Fig.~\ref{fig:methods_decoding} B,C for visualization of decoding pipeline).  
The classification pipeline is composed of logistic regression model with L1 regularization to obtain sparse weight coefficients allowing for localization of the discriminability to non-zero coefficient channels. 
Area Under Receiver Operating Characteristics (AUC) is reported as evaluation metric. All experiments contain 2 classes in classification, and the chance level of AUC is 0.5.
\par 
To complement the temporal decoding analysis and examine the spatial distribution of features contributing to classification, we computed Channel Importance Maps. These maps quantify the relative contribution of each EEG channel by calculating the proportion of timepoints—restricted to statistically significant decoding intervals—at which the channel was assigned a non-zero weight in an L1-regularized logistic regression model. Due to the sparsity-inducing property of L1 regularization, a non-zero weight indicates that a given channel contributed to the classification decision at that timepoint. Higher proportions reflect greater importance of the corresponding channel in distinguishing between groups.
\subsection{Deep Learning Model Architecture} \label{dl_arch}
The model architecture is a composition of 2 convolution layers and 2 vanilla transformer \cite{vaswani2017attention} layers with sinusoidal positional encoding and classification head. 
Adam optimizer (lr=1e-5) with a cross entropy loss and entropy regularization were used for all experiments.
For all classification experiments, batch size 512 with a five-fold cross validation stratified for gender were used. 
All models were trained for 5 runs across each fold with differing random seeds for weight initializations. 
For all experiments with class imbalance (C $vs$ D,S), minority class over-sampling was done during training and the input trials were resampled to 200 Hz.
\par
In order to reduce the intra-subject variability within brain responses to trials corresponding to a particular grouping of sentences (i.e., positive sentiment sentences or agree sentences), trials of each subject were bootstrapped.
To evaluate subject-level classification performance, we used a bootstrapping strategy detailed in Algorithm.~\ref{alg:alg1}. For each subject, B = 20 single-trial EEG responses were sampled with replacement and averaged across the time dimension to generate a single bootstrapped trial. This procedure was repeated N = 200 times, resulting in 200 bootstrapped trials per subject.
These bootstrapped trials from all training participants were used to train a classifier to predict the mental health group (e.g., depressed vs. control). The optimal model for a given experimental condition was selected using a validation set drawn from within the training data.
During testing, the classifier produced one prediction per bootstrapped trial, yielding 200 predictions per test subject. The subject-level probability of being classified as depressed was computed as the ratio of trials predicted as “depressed” to the total number of bootstrapped trials (i.e., 200). A higher probability indicates a greater likelihood of the subject belonging to the depressed group. The final classification performance was measured using Area Under the Receiver Operating Characteristic curve (AUC).
\begin{algorithm}
\caption{Bootstrapped Subject-Level EEG Classification}
\begin{algorithmic}[1]
\Require EEG trials per subject $\mathcal{T}_s \in \mathbb{R}^{N_t \times T \times C}$ where $N_t$ is number of trials, $T$ timepoints, $C$ channels
\State Set number of bootstrap samples $N \gets 200$
\State Set trials per sample $B \gets 20$
\For{each training subject $s$}
    \For{$i = 1$ to $N$}
        \State Sample $B$ trials with replacement from $\mathcal{T}_s$: $\{\mathbf{x}_1, \dots, \mathbf{x}_B\}$
        \State Compute averaged trial: $\bar{\mathbf{x}}_i \gets \frac{1}{B} \sum_{j=1}^{B} \mathbf{x}_j$
        \State Store training pair $(\bar{\mathbf{x}}_i, y_s)$
    \EndFor
\EndFor
\State Train classifier $\mathcal{M}$ on all bootstrapped training pairs
\State Select best model via validation split from training data
\For{each test subject $s'$}
    \For{$i = 1$ to $N$}
        \State Generate $\bar{\mathbf{x}}_i$ via bootstrapping as above
        \State Predict: $\hat{y}_i \gets \mathcal{M}(\bar{\mathbf{x}}_i)$
    \EndFor
    \State Compute subject probability: $P_{s'} \gets \frac{1}{N} \sum_{i=1}^{N} \mathbb{I}[\hat{y}_i = \text{depressed}]$
\EndFor
\State Compute AUC over all test participants using $P_{s'}$ and ground-truth labels
\end{algorithmic}
\label{alg:alg1}
\end{algorithm}
\subsection{Statistical Analysis}
During EEG decoding, to determine the temporal clusters in which the decoding accuracy was significantly better than chance, a non-parametric cluster-based permutation test was performed. 
Permutation test was used to compute the p-values \cite{maris2007nonparametric}. 
This was done by generating a null distribution through flipping the sign of the AUC time series at random and averaging them to generate permutations 1000 times. 
This distribution was used to convert the mean averages of the runs to a p-value series over time. 
The significance threshold was $p\leq0.05$, and the significant regions were determined using a cluster-based approach (cluster defining threshold $p\leq0.01$ and cluster size $>$ 40 ms). 
For deep learning models, 95\% Confidence intervals are reported. 
To compute these intervals, permutation of predictions were generated through sampling with replacement; this process is repeated 1000 times. 
The AUC corresponding to each permutation is computed, and the 2.5 percentile and 97.5 percentile in these 1000 values correspond to the 95\% CI intervals. 
Correlation values between model predictions and questionnaire scores when reported correspond to Spearmann correlation. Statistical differences between two correlation values were computed using a permutation test similar to comparing two AUCs.
To test whether the mean predictions between two groups of participants were statistically different, a simple t-test was performed with multiple tests using Bonferroni correction.
 
 \section*{Data Availability}
Anonymized annotations and EEG data recordings used for analysis in this paper are available upon request to shri@usc.edu.
\section*{Code Availability}
Code for temporal decoding and deep learning models is available upon request. Python (version $\geq$ 3.10), PyTorch and Scikit-learn libraries were used for model building, and Seaborn was used for visualization. Deep learning model training requires a GPU with a VRAM $\geq$ 8GB.

\section*{Acknowledgements}
This study was sponsored by the Defense Advanced Research Projects Agency (DARPA) under cooperative agreement No. N660012324006. The content of the information does not necessarily reflect the position or the policy of the Government, and no official endorsement should be inferred.
\section*{Author Contributions}
A.K. conceived the computational framework for EEG, performed the data analysis and wrote the first draft. 
W.J. pre-processed EEG recordings. 
C.M., M.H., T.M. and W.J. contributed to data acquisition in EEG recordings. 
D.B., I.B. and Th.M. contributed to stimuli design.
D.B., I.B., A.H., R.C., T.M., K.L., R.L. and S.N. developed the experimental protocol. 
A.K., W.J., K.A., S.K., A.H., R.C., R.L., E.K., D.B. and T.M. contributed to interpretation of results.
R.L. and S.N. provided oversight and supervision during the work.
All authors reviewed and approved the final manuscript.
\section*{Competing Interests}
The authors declare no competing interests.

\bibliography{main}
\newpage
\begin{appendix}
\section{D $vs$ S Decoding Results}
\begin{figure}
    \centering
    \includegraphics[width=\linewidth]{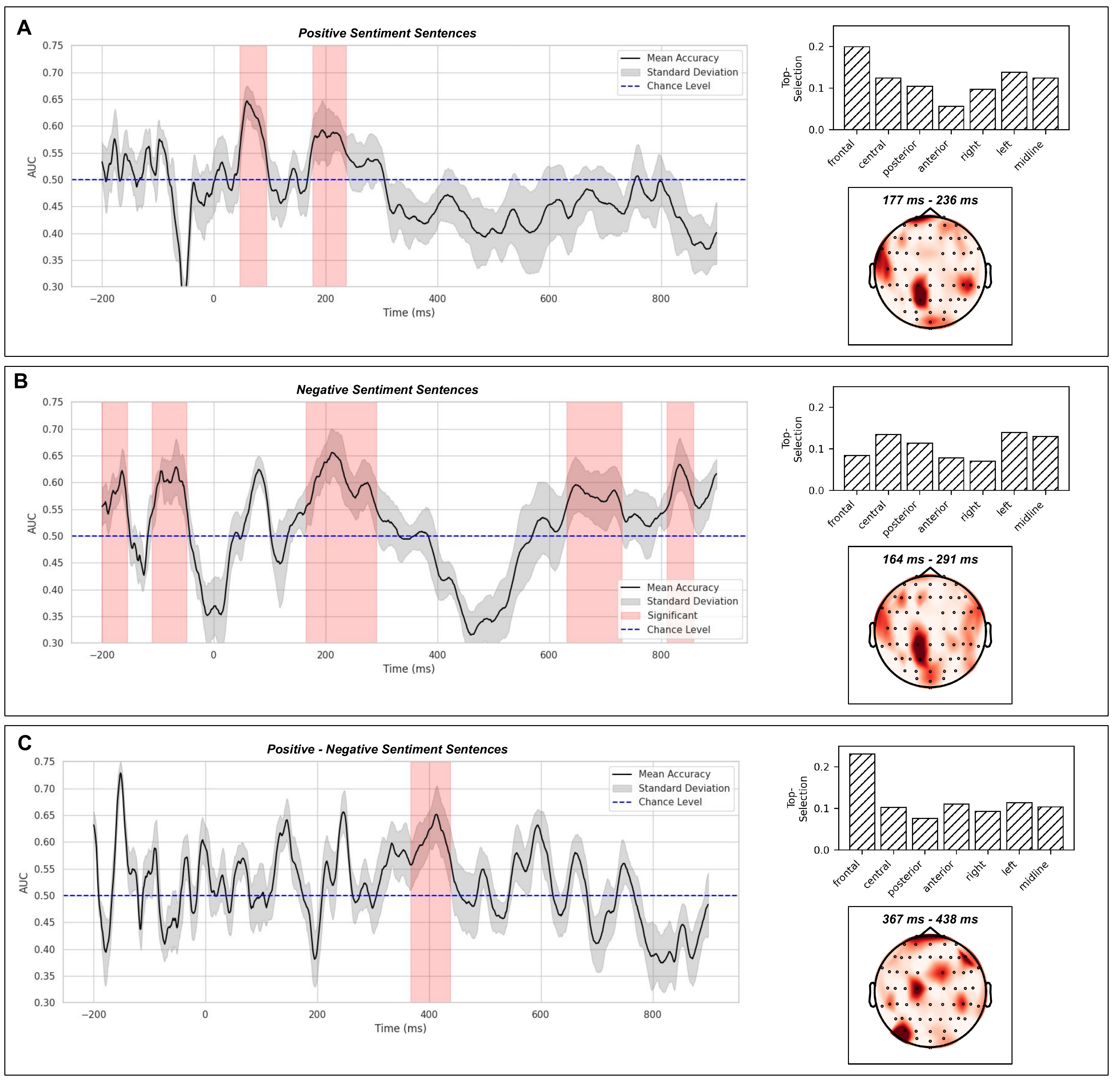}
    \caption{Temporal plots for per-timestep average AUC for Depressed $vs$ Suicidal [D $vs$ S] classification in a LOSOCV setting across 10 different seeds for (A) Positive Sentiment Sentences (B) Negative sentiment sentences (C) Positive-Negative sentiment sentences. Shaded regions in the temporal AUC plots correspond to clusters wherein the performance were found to be statistically significantly above chance from a cluster based permutation test. The histogram plot and channel importance map of the largest cluster in each setting are reported. Channel Importance Map: Each value represents the proportion of significant timepoints during which the corresponding electrode was assigned a non-zero weight in L1-regularized logistic regression, indicating its contribution to group classification. Darker regions in the weight proportion map indicate the specific region to have higher contribution to D $vs$ S classification.}
    \label{fig:ds_decoding}
\end{figure}

Temporal decoding plots for DvsS classification are reported in Fig.~\ref{fig:ds_decoding}. Negative sentiment sentences provide larger mean AUC values and larger clusters of significant decodability compared to positive sentiment sentences, which is also reflected by better performance of negative sentiment sentences in classification tasks.
In positive sentiment sentences and contrasting conditions, a frontal channel importance is observed.
This could explain the worser performance of DvsS classification models as frontal channels are often noisy leading to a lower differentiability between groups. 
\section{Ablation Experiments}
\subsection{Bootstrap Hyperparameters}
\begin{table}[]
\centering
\caption{Mean AUC and 95\% confidence intervals for sentence sentiment and response type contrasting condition classification across different number of generated trials per subject for CvsDS. In all experiments 20 trials were averaged to generate one bootstrap.}
\begin{tabular}{lcc}
\toprule
\textbf{Trials Generated (N)} & \textbf{Sentence Sentiment} & \textbf{Response Type} \\
\midrule
\textbf{100}  & 0.696 (0.651, 0.738) & 0.661 (0.618, 0.702) \\
\textbf{200}  & 0.707 (0.666, 0.749) & 0.688 (0.644, 0.728) \\
\textbf{400}  & 0.699 (0.653, 0.740) & 0.668 (0.625, 0.709) \\
\textbf{1000} & 0.678 (0.634, 0.718) & 0.625 (0.584, 0.669) \\
\bottomrule
\end{tabular}
\label{tab:gen_trials_ci}
\end{table}

\begin{table}[]
\centering
\caption{Mean AUC and 95\% confidence intervals across number of trials averaged per bootstrap trial generated per subject for sentence sentiment and response type contrasting condition classification for CvsDS. For all experiments 200 trials per subject were generated.}
\begin{tabular}{lcc}
\toprule
\textbf{Trials Mean (B)} & \textbf{Sentence Sentiment} & \textbf{Response Type} \\
\midrule
\textbf{5}  & 0.670 (0.626, 0.712) & 0.641 (0.598, 0.682) \\
\textbf{10} & 0.681 (0.635, 0.722) & 0.661 (0.618, 0.703) \\
\textbf{20} & 0.707 (0.666, 0.749) & 0.688 (0.644, 0.728) \\
\textbf{40} & 0.699 (0.656, 0.741) & 0.661 (0.618, 0.703) \\
\bottomrule
\end{tabular}
\label{tab:trialwise_perf}
\end{table}

Bootstrapped EEG trials were used as inputs to the EEG models. 
Tab.~\ref{tab:gen_trials_ci} shows the ablations of the number of bootstrap trials generated and Tab.~\ref{tab:trialwise_perf} indicates the number of EEG trials averaged per bootstrap trial. For the number of bootstrap trials generated, no significant difference is observed when varying the number of bootstrap trials from 100 per subject to 1000 per subject. 
However, as the number of averaged trials per bootstrap increases from 5 to 20, a relative improvement of 5\% is observed. 
Further increase in averaged trials results in a reduction in AUC, which could be owing to reduced variation from the larger number of sampled trials.
\subsection{Neutral Sentiment Sentences for Classification}
\begin{table}[]
\centering
\caption{Mean AUC and 95\% confidence intervals for classification across different group comparisons (Control $vs$ Depressed+Suicidal and Depressed $vs$ Suicidal) when using Neutral sentiment sentences and contrasting with Neutral sentiment sentences.}
\begin{tabular}{lccc}
\toprule
\textbf{C $vs$ DS} & \textbf{Positive-Neutral} & \textbf{Neutral} & \textbf{Negative-Neutral} \\
\midrule
\textbf{AUC} & 0.634 & 0.617 & 0.641 \\
95\% CI & (0.590, 0.679) & (0.575, 0.660) & (0.598, 0.681) \\
\bottomrule
\end{tabular}

\vspace{0.5cm}

\begin{tabular}{lccc}
\toprule
\textbf{D $vs$ S} & \textbf{Positive-Neutral} & \textbf{Neutral} & \textbf{Negative-Neutral} \\
\midrule
\textbf{AUC} & 0.597 & 0.529 & 0.634 \\
95\% CI & (0.543, 0.654) & (0.475, 0.588) & (0.581, 0.689) \\
\bottomrule
\end{tabular}
\label{tab:groupwise_sentiment}
\end{table}
In addition to positive and negative sentiment sentences, Table~\ref{tab:groupwise_sentiment} shows the results for employing neutral sentiment sentences for classification of C $vs$ DS and D $vs$ S participants.
When considering neutral sentences singularly, we observe that the performance is lower compared to positive sentences and is higher compared to negative sentences, which is akin to the observations in decoding experiments.
Interestingly, we observe that for D $vs$ S contrasting negative and neutral sentences provides a mean AUC=0.634, which is higher than contrasting positive and negative sentiment sentences.
This provides promise for exploration into more sophisticated methods of fusion.
\subsection{CD $vs$ S classification}
Table~\ref{tab:cd_s} provides classification results for CD $vs$ S classification. Both the sentence sentiment and response type results indicate that models are unable to distinguish suicidal participants from control and depressed effectively. This could be owing to the responses of depressed and suicidal participants being similar, which does not enable the models to differentiate suicidal participants. 
\begin{table}[]
\centering
\caption{Mean AUC and 95\% confidence intervals for CD $vs$ S classification based on sentence sentiment and response type contrasts.}
\begin{tabular}{lccc}
\toprule
\textbf{Sen Sent} & \textbf{Positive-Negative} & \textbf{Positive} & \textbf{Negative} \\
\midrule
 AUC & 0.593 & 0.565 & 0.559 \\
95\% CI    & (0.553, 0.636) & (0.518, 0.612) & (0.513, 0.607) \\
\midrule
\textbf{Response Type} & \textbf{Agree-Disagree} & \textbf{Agree} & \textbf{Disagree} \\
\midrule
 AUC & 0.520 & 0.572 & 0.564 \\
95\% CI    & (0.478, 0.567) & (0.527, 0.619) & (0.514, 0.610) \\
\bottomrule
\end{tabular}
\label{tab:cd_s}
\end{table}

\section{Detail Demographic Information}
\begin{table}[h!]
\centering
\begin{tabular}{lrrrrrr}
\toprule
 & \multicolumn{2}{c}{\textbf{Control}} & \multicolumn{2}{c}{\textbf{Depressed}} & \multicolumn{2}{c}{\textbf{Suicidal}} \\
\cmidrule(lr){2-3} \cmidrule(lr){4-5} \cmidrule(lr){6-7}
\textbf{Demographic Attribute} & \textbf{N} & \textbf{\%} &
\textbf{N} & \textbf{\%} & 
\textbf{N} & \textbf{\%} \\
\midrule
\textit{Gender} & & & & & & \\
\hspace{5mm}Female & 30 & 61.2 & 28 & 59.6 & 33 & 66.0 \\
\hspace{5mm}Male & 17 & 34.7 & 13 & 27.7 & 15 & 30.0 \\
\hspace{5mm}Other & 2 & 4.1 & 6 & 12.8 & 2 & 4.0 \\

\textit{Ethnicity} & & & & & & \\
\hspace{5mm}Asian & 26 & 53.1 & 21 & 44.7 & 20 & 40.0 \\
\hspace{5mm}Black & 1 & 2.0 & 5 & 10.6 & 8 & 16.0 \\
\hspace{5mm}Hispanic/Latinx & 9 & 18.4 & 8 & 17.0 & 15 & 30.0 \\
\hspace{5mm}Middle Eastern & 2 & 4.1 & 3 & 6.4 & 2 & 4.0 \\
\hspace{5mm}Native American/Alaskan Native & 0 & 0.0 & 1 & 2.1 & 0 & 0.0 \\
\hspace{5mm}White (Non-Hispanic) & 13 & 26.5 & 16 & 34.0 & 13 & 26.0 \\

\textit{Student status} & & & & & & \\
\hspace{5mm}Freshman & 8 & 16.3 & 8 & 17.0 & 10 & 20.0 \\
\hspace{5mm}Sophomore & 5 & 10.2 & 4 & 8.5 & 12 & 24.0 \\
\hspace{5mm}Junior & 15 & 30.6 & 13 & 27.7 & 9 & 18.0 \\
\hspace{5mm}Senior & 7 & 14.3 & 7 & 14.9 & 13 & 26.0 \\
\hspace{5mm}Graduate & 12 & 24.5 & 12 & 25.5 & 5 & 10.0 \\
\hspace{5mm}Other & 2 & 4.1 & 1 & 2.1 & 1 & 2.0 \\
\bottomrule
\end{tabular}
\caption{\textbf{Demographics of the data collection} across the 146 participants that were used for analysis. The table reports the distribution of participants by demographic attributes that include gender, ethnicity, and student status. Values are presented as counts and percentages within each group. Age is not reported; participants were required to be between 18 and 25 years old.}
\end{table}
    
\end{appendix}

\end{document}